\documentclass[conference]{IEEEtran}
\IEEEoverridecommandlockouts
\usepackage{cite}
\usepackage{amsmath,amssymb,amsfonts}
\usepackage{algorithmic}
\usepackage{graphicx}
\usepackage{textcomp}
\usepackage{xcolor}
\usepackage{paralist}
\def\BibTeX{{\rm B\kern-.05em{\sc i\kern-.025em b}\kern-.08em
    T\kern-.1667em\lower.7ex\hbox{E}\kern-.125emX}}
\begin{document}

\title{Kick-motion Training with DQN in AI Soccer Environment\\
\thanks{This work was supported by the Institute for Information communications Technology Promotion (IITP) grant funded by the Korean government (MSIT) (No.2020-0-00440, Development of Artificial Intelligence Technology that continuously improves itself as the situation changes in the real world).}
}

\author{\IEEEauthorblockN{Bumgeun Park}
\IEEEauthorblockA{\textit{Cho Chun Shik Graduate School of Mobility} \\ 
\textit{Korea Advanced Institute of Science and Technology (KAIST)}\\
Daejeon, South Korea\\
j4t123@kaist.ac.kr}
\and
\IEEEauthorblockN{Jihui Lee}
\IEEEauthorblockA{\textit{Division of Future Vehicle} \\
\textit{Korea Advanced Institute of Science and Technology (KAIST)}\\
Daejeon, South Korea \\
jihui@kaist.ac.kr}
\and
\IEEEauthorblockN{Taeyoung Kim}
\IEEEauthorblockA{\textit{Cho Chun Shik Graduate School of Mobility} \\
\textit{Korea Advanced Institute of Science and Technology (KAIST)}\\
Daejeon, South Korea \\
ngng9957@kaist.ac.kr}
\and
\IEEEauthorblockN{Dongsoo Har}
\IEEEauthorblockA{\textit{Cho Chun Shik Graduate School of Mobility} \\
\textit{Korea Advanced Institute of Science and Technology (KAIST)}\\
Daejeon, South Korea \\
dshar@kaist.ac.kr}
}

\maketitle

\begin{abstract}
This paper presents a technique to train a robot to perform kick-motion in AI soccer by using reinforcement learning (RL). In RL, an agent interacts with an environment and learns to choose an action in a state at each step. When training RL algorithms, a problem called the curse of dimensionality (COD) can occur if the dimension of the state is high and the number of training data is low. The COD often causes degraded performance of RL models.
In the situation of the robot kicking the ball, as the ball approaches the robot, the robot chooses the action based on the information obtained from the soccer field. In order not to suffer COD, the training data, which are experiences in the case of RL, should be collected evenly from all areas of the soccer field over (theoretically infinite) time.
In this paper, we attempt to use the relative coordinate system (RCS) as the state for training kick-motion of robot agent, instead of using the absolute coordinate system (ACS). Using the RCS eliminates the necessity for the agent to know all the (state) information of entire soccer field and reduces the dimension of the state that the agent needs to know to perform kick-motion, and consequently alleviates COD.
The training based on the RCS is performed with the widely used Deep Q-network (DQN) and tested in the AI Soccer environment implemented with Webots simulation software.
\end{abstract}

\begin{IEEEkeywords}
Reinforcement learning (RL), Deep Q-Network (DQN), AI soccer, Curse of dimensionality (COD), Coordinate transformation matrix (CTM)\\
\end{IEEEkeywords}

\section{Introduction}
Reinforcement Learning (RL), which is one of the widely used machine learning techniques, makes an agent maximize the accumulated rewards in given environment \cite{rl}. To do so requires the use of experiences, e.g., experience replay, during training. The agent selects an action based on the current state and receives the reward from the environment. 
Through this interaction, the agent can learn which action is more useful to achieve the goal.
Recently, a deep neural network (DNN) for nonlinear approximation enables the RL to handle more complicated tasks such as playing a range of Atari games \cite{atari}, playing the game of GO \cite{go1,go2}, controlling the robot arm \cite{robot_arm1,robot_arm2,robot_arm3}, and planning path for mobile robots \cite{path_planning1,path_planning2}.

AI soccer environment, which is one of the environments for training multi-agent RL algorithms, is introduced in the previous work \cite{aisoccer2}.
The main target of AI Soccer is winning the soccer game by controlling each robot. 
Each robot's movement consists of many small motions such as "going to a target spot", "blocking robot or ball", "positioning", "heading", and "kicking". Therefore, the entire AI soccer training can be divided into various tasks of training motions. The state, action, and reward, which are three basic elements for RL, should be designed depending on the motion to be trained.

The curse of dimensionality(COD) occurs when the dimension of the state increases and the amount of training data is inappropriately small. The inappropriately small data for training causes the RL models to be overfitted and consequently leads to deteriorated performance.


In this paper, we consider training kick-motion with a deep Q-network (DQN).
To train kick-motion successfully, the state should include various information about the robot and ball on the soccer field, which is described by their current position, speed, and the direction of the movement of ball or robot. However, designing the state to include large information can cause the COD, unless experiences are gained evenly over all areas of the soccer field. To eliminate the need for the agent to know all the game information over entire soccer field and to reduce the dimension of the state, we attempt to design the state for efficient kick-motion training by using a coordinate transformation matrix (CTM) used to transform coordinates in robotics or computer vision \cite{rotation}. The main contributions of this paper are as follows:
\begin{enumerate}
\item The attempted designing of the state avoids deterioration arising from the COD and improve the performance of the RL algorithm.
\item The RCS taken for AI Soccer eliminates the need for excessive exploration and reduces the dimension of the state.
\end{enumerate}

\section{Background}
\subsection{AI soccer environment}
In this paper, experiments are conducted on the AI soccer environment employed for AI World Cup event \cite{aisoccer1}. The environment consists of a ball, two teams of robots and a soccer field. Each team consists of five robots each driven by two wheels. Like the human soccer game, each team tries to beat the other team by controlling their robots to take the best action for each game state. The robots are controlled at each timestep by setting wheel speeds. The two separately controllable wheels extend the range of actions and increase the degree of freedom. Value of the best action can be measured by corresponding reward.
Each robot, as an agent, takes an action based on the situation of the field to get a higher accumulated reward. This is a very challenging task, because various strategies are needed for the agents to adapt to the vast state space which is observed in dynamic game situation. 

\subsection{Curse of dimensionality (COD)}
Traditional RL algorithms suffer from the COD when the dimension of the space in the environment is excessively high \cite{curse_of_dimensionality}. As the dimension of the state increases, the number of experiences needed to train an agent grows exponentially. The lack of experiences makes the RL model overfitted. Although using function approximation such as DNN alleviates this problem, the performance of the algorithm is still deteriorated.
In the AI soccer environment, large information describing the (game) state and two-wheeled actions with a high degree of freedom increase the dimension of the state space and action space, which readily causes the COD.

\subsection{Deep Q-network (DQN)}
The DQN algorithm, which is a widely used RL algorithm, is a deep RL method that approximates an action-value function, called Q-function by using a DNN.
The Q-function with parameters $\theta$ is defined as follows:
\begin{equation}Q^{\theta}(s,a)=\mathbb{E}_{s,a,\theta}[\sum_{t=1}^{\infty}\gamma^{t-1} r_{t}]\end{equation}
where a, s, r, and $\gamma$ are action, state, reward, and discount factor respectively.
DQN is trained to minimize the loss function, which can be presented as follows:
\begin{equation}L(\theta)=\mathbb{E}[(y_{t}-Q^{\theta}(s_{t},a_{t}))^2]\end{equation}
The target Q function $y_{t}$ is calculated by reward and next state $s'$ as follows:
\begin{equation}y_t=r(s_t,a_t)+\gamma max_{a'}Q^{\theta^{-}}(s',a')\end{equation}
where $\theta^{-}$ represents the parameters of a fixed and separate target network.\\

\begin{figure}[t!]
\centerline{\includegraphics[width=0.45\textwidth]{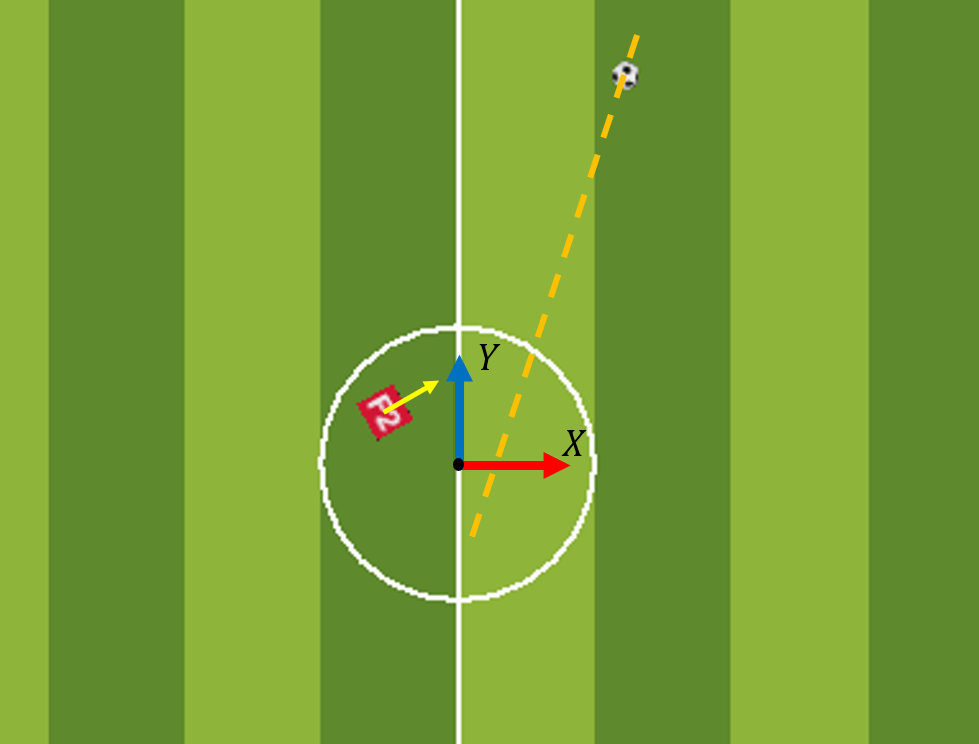}}
\caption{Illustration of the absolute coordinate system (ACS). The origin of the ACS is placed in the center of the field. The orange dotted line presents the direction of ball movement. The yellow arrow presents the direction the robot movement.}
\label{ACS}
\end{figure}

\begin{figure}[t!]
\centerline{\includegraphics[width=0.45\textwidth]{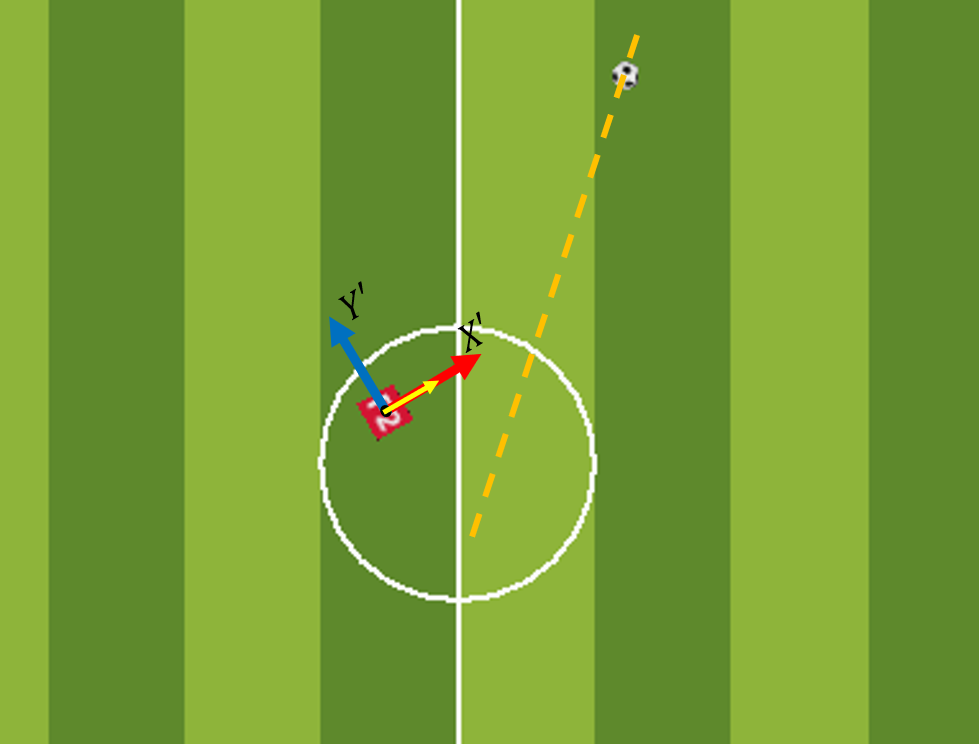}}
\caption{Illustration of the relative coordinate system (RCS). The origin of the RCS is placed in the center of the robot, and the direction the robot is aligned with the $X'$-axis.}
\label{RCS}
\end{figure}

\section{Relative coordinate system for description of game state}
In the task of training the robot to perform kick-motion, the robot chooses at each timestep the best action out of a set of actions, taking into account the information about the ball and the robot such as the current coordinates, the velocity of the ball and the robot, and the approaching angle of the ball to the robot.
To avoid the COD and train the robot to kick the ball at any location, it is required to use evenly distributed experiences over entire field.
For this purpose, we adopt the relative coordinate system (RCS) to represent the state, instead of the absolute coordinate system (ACS). The RCS can alleviate the COD and make training process straightforward. Most important benefit obtained from the use of the RCS is that the robot can achieve the task at any location after training at only one location.

In the AI soccer environment, all raw information given to the robot from the environment is presented in the ACS format. The origin of the ACS is defined as the center of the field with $X$-axis and $Y$-axis as shown in Fig.~\ref{ACS}. In this case, the state given to the robot is defined as follows:
\begin{equation}
S_{ACS}=[R_x,R_y,R_\theta,B_x,B_y,V_{R_x},V_{R_y},V_{B_x},V_{B_y},\theta]\label{eq4}
\end{equation}
Each element in \eqref{eq4} respectively represents $X$ and $Y$ coordinates of the robot, the heading angle of the robot, $X$ and $Y$ coordinates of the ball, $X$ and $Y$ direction velocity of the robot, $X$ and $Y$ direction velocity of the ball, and the angle of the direction in which the ball moves.
The high dimensional state space formulated in \eqref{eq4} can deteriorate the performance of RL algorithms.

With the RCS, the dimensional space is reduced and training becomes efficient.
In the RCS, the origin is defined at the center of the robot with $X'$-axis and $Y'$-axis and the direction the robot movement is the same as the direction of $X'$-axis shown in Fig.~\ref{RCS}.
With the RCS, the values of $R_x$, $R_y$, and $R_\theta$ are always zero since the origin of the RCS is the center of the robot. The value of $V_R{}_y$ is also always zero because the robot can take only one of forward, backward, or stop, during the task of the kick-motion.
As a result, the state can be defined as follows:
\begin{equation}
S_{RCS}=[B_x^{RCS},B_y^{RCS},V_{R_x}^{RCS},V_{B_x}^{RCS},V_{B_y}^{RCS},\theta^{RCS}]\label{eq}
\end{equation}
where $(\cdot)^{RCS}$ means $(\cdot)$ in the RCS.
The relationship between the RCS and the ACS can be presented using the CTM as shown in \eqref{CTM}.

\begin{equation}
M
=
\begin{bmatrix}
\cos(R_\theta) & -\sin(R_\theta) & R_x\\
\sin(R_\theta) & \cos(R_\theta) & R_y\\
0 & 0 & 1
\end{bmatrix}
\label{CTM}
\end{equation}
The position, velocity, and angle in the RCS can be computed with the CTM and those in the ACS, and can be shown as \eqref{RCS_ACS}, \eqref{velocity}, and \eqref{angle}.

\begin{equation}
\begin{bmatrix}
B_x^{RCS} & V_{B_x}^{RCS}\\
B_y^{RCS} & V_{B_y}^{RCS}\\
1 & 1
\end{bmatrix}=
M^{-1}\begin{bmatrix}
B_x & V_{B_x}\\
B_y & V_{B_y}\\
1 & 1
\end{bmatrix}\label{RCS_ACS}
\end{equation}

\begin{equation}
V_{R_x}^{RCS}={V_{R_x}}^2+{V_{R_y}}^2\label{velocity}
\end{equation}

\begin{equation}
\theta^{RCS}=\theta-R_{\theta}\label{angle}
\end{equation}

The change of the coordinate system from the ACS to the RCS reduces the dimension of the state by 60\%, which means significantly easier training process.\\

\begin{figure}[t!]
\centerline{\includegraphics[width=0.45\textwidth]{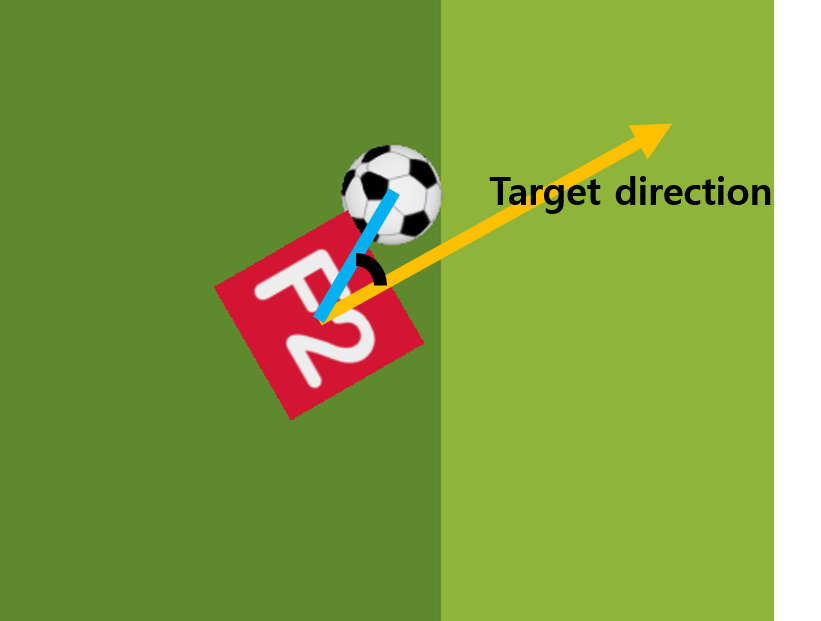}}
\caption{Illustration of a situation in which the ball and the robot are in contact. The blue line connects from the center of the robot to the center of the ball and the yellow arrow presents the direction toward the target point.}
\label{contact}
\end{figure}

\section{Experiments}
This section describes AI Soccer environment and presents the result of experiments conducted to show the effect of the COD.

\subsection{Experimental Environment}
Experiments are conducted for the kick-motion task in the AI soccer environment developed with Webot physics simulator \cite{webots}. The goal of the kick-motion task is that robot kicks the ball to a target point. The robot is facing the target point and the ball is located at a random point with a distance of 2m from the robot. With the start of an episode of training, the ball starts to move at a random speed to a random point near the robot. At each timestep, the robot chooses an action based on the state. Each episode consist of 40 experiences over 40 timesteps. The RL framework for this task can be described as follows:
\begin{compactitem}
    \item[\textbf{States:}] State consists of 6 values shown as \eqref{eq} in Section III.
    \item[\textbf{Rewards:}] Reward is given as a sum of 3 types of reward, $r_{contact} + r_{theta} + r_{velocity}$. The $r_{contact}$ is a reward of contact, which is -0.1 before the robot contact the ball and 0 after the contact. The $r_{Theta}$ is a reward according to distance of the contact point from the center of the robot, which is represented as $3 \cos{\theta_{contact}}$, where $\theta_{contact}$ is the angle between the blue and the yellow lines in Fig.~\ref{contact}. The $r_{velocity}$ is a reward according to the speed of the contact with the ball, which is represented as $3.92 V_{contact}$, where $V_{contact}$ is the velocity of the robot at the contact.
    \item[\textbf{Actions:}] List of available actions: going straight forward with steady-state speeds of 2.55 (m/s), stop, going straight backward with steady-state speeds of -2.55 (m/s).
\end{compactitem}

\subsection{Experimental Results}
The performances with the kick-motion task are evaluated in terms of the sum of the reward for each training episode.
To reduce granularity of the reward variation, the moving average of the sum of the reward across 250 episodes is taken.

\begin{figure}[t!]
\centerline{\includegraphics[width=0.5\textwidth]{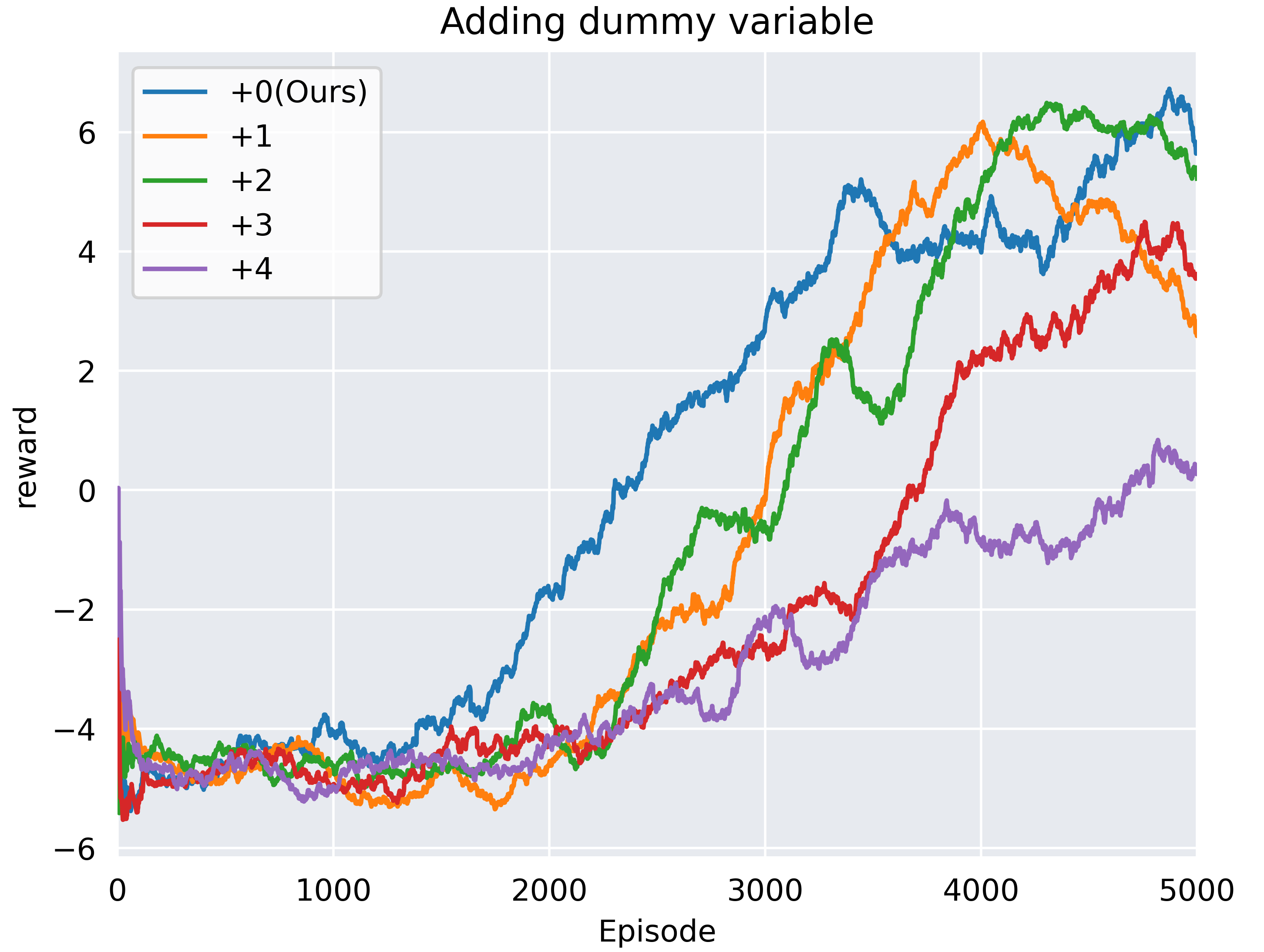}}
\caption{Total reward according to the number of dummy variables. It shows that increasing the dimension of state space slows down the training speed and as a result degrades performance. The blue line represents the model with the RCS and the others represent comparative models. Especially, the purple line is equivalent to the model with the ACS.}
\label{reward}
\end{figure}

To present the intuitive understanding of the COD, models used to compare with the proposed method are trained with states including 'dummy variables', which are some meaningless random variables. The states for the comparison models are the state in the RCS with 'dummy variables'. In Fig.~\ref{reward}, the name of '$+n$' means that $n$ dummy variables are added to the state.

Figure ~\ref{reward} shows the results of the experiments. The (total accumulated) reward of the action for RCS based state exceeds -4 from 1500-th episode, which is about 500 episodes faster than by ACS based method. The RCS based method achieves the highest reward value of 6.8, although the reward value of the RCS based method is similar to the others in the latter part of the training, e.g., 3600-4700 episodes.

\section{Conclusion}
Efficient training of the RL algorithms with high dimensional state space is hard to achieve, due to the COD.
In this paper, a method to reduce the dimension of the state space for training the kick-motion task in the AI soccer environment is attempted.
The attempted method uses the CTM to convert the state in the ACS to the state in the RCS. The dimension of the state space required for the agent to achieve the kick-motion is reduced from ten to six since 4 values in the state are always zero by the RCS based method.
The experimental result shows that the kick-motion can be achieved with the reduced state by the RCS based method. The result obtained along with the comparative models with different number of variables, which are trained with the states including dummy variables, shows that the lower dimension of the state space can avoid deterioration due to the COD.

\section*{Acknowledgment}

This work was supported by the Institute for Information communications Technology Promotion (IITP) grant funded by the Korean government (MSIT) (No.2020-0-00440, Development of Artificial Intelligence Technology that continuously improves itself as the situation changes in the real world).\\

\bibliographystyle{unsrt}
\bibliography{reference}
\vspace{12pt}

\end{document}